\documentclass[12pt]{article}
\usepackage{scicite}
\usepackage[none]{hyphenat}
\usepackage{cite}
\usepackage{times}
\usepackage{graphicx}
\usepackage{subfig}
\usepackage{mathtools}
\usepackage{tikz}
\usepackage{pgfplots}
\usepackage{smartdiagram}
\usesmartdiagramlibrary{additions}
\usepackage{hyperref}
\hypersetup{
    colorlinks=true,
    linkcolor=blue,
    filecolor=magenta,      
    urlcolor=cyan,
}
\usepackage{float}
\usepackage{cleveref}
\usepackage[document]{ragged2e}
\restylefloat{table}
\newcounter{lastnote}

\title{FUNN: Flexible Unsupervised Neural Network} 
\author
{David Vigouroux, Sylvain Picard\\
\\
\normalsize{IRT Saint Exupery, Toulouse, France}}

\date{}

\begin{document}

\baselineskip24pt

% Make the title.

\maketitle

\justify
\begin{abstract}
Deep neural networks have demonstrated high accuracy in image classification tasks. However, they were shown to be weak against adversarial examples: a small perturbation in the image which changes the classification output dramatically. In recent years, several defenses have been proposed to solve this issue in supervised classification tasks. We propose a method to obtain robust features in unsupervised learning tasks against adversarial attacks. Our method differs from existing solutions by directly learning the robust features without the need to project the adversarial examples in the original examples distribution space. A first auto-encoder A1 is in charge of perturbing the input image to fool another auto-encoder A2 which is in charge of regenerating the original image. A1 tries to find the less perturbed image under the constraint that the error in the output of A2 should be at least equal to a threshold. Thanks to this training, the encoder of A2 will be robust against adversarial attacks and could be used in different tasks like classification. Using state-of-art network architectures, we demonstrate the robustness of the features obtained thanks to this method in classification tasks.
\end{abstract}

\section{Introduction}
\paragraph{}
Neural networks and particuraly convolutional neural networks have shown impressive results in many different tasks in computer vision such as object detection, image recognition and segmentation. A down side of these networks is their lack of robustness as described in  \cite{goodfellow_explaining_2014}, \cite{kurakin_adversarial_2016-1}, \cite{szegedy_intriguing_2013}.
Indeed, imperceptible perturbations of the input (image, sound…) can confuse the network and completely change the prediction. This kind of input is described as an "Adversarial Example". This misclassification could threaten the security and the safety of critical systems. By using the concept of adversarial examples it is possible to design algorithms in order to generate inputs that will modify the prediction with a high rate of succes and a very little perturbation. Those algorithms are known as "Adversarial Attacks". We commonly distinguish two types of attacks. White box attacks: when the network architecture as well as its weights are known. Black box attacks: when they are not known by the attacker. Our work is focused on  untargeted  white-box  attacks computed  using  different attack methods.

The field of adversarial generation is an active research field, many methods have been developped in order to generate adversarial example as described in \cite{yuan_adversarial_2017}. 
In this work we use five of the most used attack methods in litterature: 
The Fast  Gradient Sign  Method  (FGSM)  \cite{goodfellow_explaining_2014}, Iterative FGSM \cite{kurakin_adversarial_2016},  Single Pixel attack and LocalSearch \cite{narodytska_simple_2016} and  Deepfool  \cite{moosavi-dezfooli_deepfool:_2015}.
Those methods are gradient-based or score-based aproaches that try to find the minimal perturbation that will modify the model prediction.
Several studies have conjectured that this lack of robustness comes from the fact that the adversarial examples are out of the datasets distribution, near the border of decision. Different methods have been developped in order to be robust against adversarial attacks.
Some strategies try to project the adversarial examples in the original distribution as in  \cite{meng_magnet:_2017}, \cite{samangouei_defense-gan:_2018}, \cite{santhanam_defending_2018}. Other approaches propose to add adversarial examples into the training dataset \cite{szegedy_intriguing_2013}, \cite{goodfellow_explaining_2014}, \cite{moosavi-dezfooli_deepfool:_2015}, similar to data augmentation. Finally new methods like \cite{munusamy_kabilan_vectordefense:_2018} transform the input data such as it is less sensible to perturbations. The problem with those methods is that they are attack-specific and then not efficient against new or simply different attacking method. Recently new methods proposed to learn the attacking et defending concept using neural networks. Some methods use Generative Adversarial Networks (GANs) to generate adversarial examples as in \cite{samangouei_defense-gan:_2018}, \cite{shen_ape-gan:_2017}, \cite{lee_generative_2017} or Auto-Encoders as in \cite{folz_adversarial_2018}, \cite{srinivasan_counterstrike:_2018}, \cite{baluja_learning_nodate}. This paper is an attempt to increase an autoencoder(A2) robustness by adversarial training, using another autoencoder(A1) used to generated perturbed images during training. This new method computes robust features without the need to add any additional networks. Only the weights of the network are modify to obtain a robust network without loss of performance.
 
\section{Related Work}
\subsection{Attack strategies}
\paragraph{}
Many attack strategies have been proposed, which can be classified in two categories: black-box and white-box attacks. White-box attacks have access to all the weights and gradients of the
classifier while Black-box strategies have access only to the predictions of the network. In this work we will focus on white-box attacks. We use two kinds of white-box attacks, gradient-based and score-based. Gradient-based methods are:

\paragraph{Fast Gradient Sign Method (FGSM)}
Given an image x and its corresponding label y, the FGSM attack sets the perturbation $\delta$ to:
\begin{equation}\label{eq:fgsm}
\delta = \epsilon \cdot sign(\nabla_{x}J(x,y))
\end{equation}
FGSM \cite{goodfellow_explaining_2014} uses the sign of the gradient at each pixel to determine the direction with which to change the corresponding pixel value.

\paragraph{DeepFool}
DeepFool \cite{moosavi-dezfooli_deepfool:_2015} perform an iterative attack with a linear approximation in order to find the closest distance from the original input to the decision boundary of adversarial examples. If $f$ is a binary differentiable classifier, they used an iterative
method to approximate the perturbation by considering $f$ is linearized around $x_{i}$ at each iteration. The minimal perturbation is computed as:
\begin{equation}\label{eq:deepfool}
\begin{multlined}
\underset{\eta_{i}}{arg min} ||\eta_{i}||_{2} \\
s.t. f(x_{i}) + \nabla f(x_{i})^{T} \eta_{i} = 0
\end{multlined} 
\end{equation}
This result can also be extended to the multi-class classifier by finding the closest hyperplanes.
DeepFool provided less perturbation and reduced intensity compared to FGSM and BIM.
\\
Score-based methods are:
\paragraph{Single Pixel}
Change the value of a random pixel to the min or max value of the image. If the perturbed image is not fooling the classifier, the pixel value is reseted to it original value and the algorithm choose another random pixel. The method is iterative and stop if an adversarial image is generated or if the max-step limit is reached.

\paragraph{Local Search} 
Use a local search procedure to find a pixel (or a group of pixels) that is critical for the classfifier robustness and then modify it value to generate adversarial images. Doing this there might exist a pixel (or a group of pixels) in the adversarial image whose coordinate value could lie outside the valid range of image values. The LocSearchAdv algorithm finds pixel locations to perturb using and then applies a defined transformation function to these selected pixels to generated an adversarial image.

\subsection{Defense strategies}
Mutiple defense methods have been used to increase robustness of deep neural networks against adversarial attacks. This section describes the most used strategies.

\paragraph{Adversarial training}
Adversarial training is an intuitive methods consisting of augmenting the training dataset with adversarial example. Thi method is efficient to increase robustness against the adversarial examples used in training. The main issue with this methods is that it is attack-agnostic and badly transfers to new adversarials examples not used in training. Additionally, it tends to make the model more robust to white-box attacks than to black-box attacks due to gradient masking.

\paragraph{Defensive distillation}
Defensive distillation \cite{papernot_distillation_2015} is a method in which a classifier is trained in two rounds using a variant of the distillation \cite{hinton_distilling_2015} method. This induce learning a smoother network and reducing the amplitude of gradients around input points, increase the robustness against adversarials. This method have been proven to be inneficient against new, black-box attacks \cite{carlini_defensive_2016}.

\paragraph{Adversarial Detectors}
Another idea of defense is to detect adversarial examples \cite{metzen_detecting_2017}. For each attack generating method considered, a deep neural network classifier (detector) is trained to tell whether an input is normal or adversarial. The detector was directly trained on both normal and adversarial examples. It showed good performance when the training and testing attack examples were generated from the same process and the perturbation was large enough, but it did not generalize well across different attack parameters and attack generation processes.

\paragraph{MagNet}
MagNet \cite{meng_magnet:_2017} is a method based on adversarial detectors strategy. It trains a reformer network (which is an auto-encoder or a collection of auto-encoders) to move adversarial examples closer to the manifold of legitimate, or natural, examples. It was proven to be an effective strategy against gray-box attacks where the attacker is aware of the network architecture and defenses but do not know it parameters.

\section{Proposed Method}
\paragraph{}
The previously described defense methods provide intuition on what can make a network robust to adversarial examples. Our strategy differs from them by doing sort of adversarial training with data generated thanks to a separated network during training. The generating network is used only in training, and not in inference unlike in the Adversarial Detectors strategies. This paper is focused only on unsupervised learning strategies, this is the reason why we decided to use Auto-Encoders. An auto-encoder is a deep neural network composed by an encoder responsible of deep features extraction from input image and a decoder responsible of image re-generation from deep features extracted by the encoder. During training the network is optimised in order to generate images similar to input imagse.
We propose a defense strategy to increase robustness against white box attacks using $L_{2}$ norm and demonstrate its efficiency against $L_{\infty}$. The strategy used is to train two Auto-Encoders, an “attacker”: a generator of adversarial examples from a given image and a “defender”: capable to extract the original image from adversarial examples given by the attacker. The attacker and the defender are trained simultaneously and adapt themselves all along of the training until convergence. At the end of training, we expect the deep features extracted by the defender's encoder to be robust against the images generated by the attacker and that they could be used in further tasks like classification. We expect this training method to generate deep features representing a larger distribution of input images and so increase the robustness of the defending auto-encoder(A2).

\begin{center}
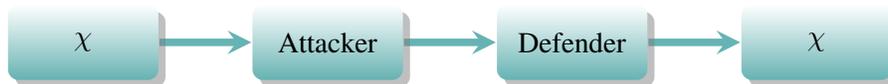

  \smartdiagramset{border color=none, uniform color list=teal!60 for 4 items, arrow style=<-, module x sep=3.25, back arrow disabled,}
  \smartdiagram[flow diagram:horizontal]{$\chi$, Attacker, Defender, $\chi$}
  \captionof{figure}{Concept Architecture}
\end{center}

The goal of the attacker A1 is to find the minimal perturbation which fools the defender A2 with a loss value at least equal to a threshold $\beta$. The defender should avoid to be deceived.

\begin{equation}\label{eq:defense}
\min_{\theta_{defense}} E_{x \sim \chi}(L_{\alpha}(x, Defense(Attack(x,\theta_{attack} ),\theta_{defense})))
\end{equation}

\begin{equation}\label{eq:attack}
\min_{\theta_{attack}} E_{x \sim \chi}(L_{\alpha}(Attack(x, \theta_{attack}), x))
\end{equation}

Under the constraint:
\begin{equation}\label{eq:constraint}
E_{x \sim \chi}(L_{\alpha}(x, Defense(Attack(x, \theta_{attack}), \theta_{defense}))) \geq \beta
\end{equation}

\paragraph{}
Where:

\setlength{\parindent}{15ex}
$\chi$ is the ensemble of distribution examples 

$L_{\alpha}$ is the loss function : $\chi, \chi \to R$

$Attack(x, \theta_{attack}): \chi \to \chi$ with parameters $\theta_{attack}$

$Defense(x, \theta_{defense}): \chi \to \chi$ with parameters $\theta_{defense}$

$\theta_{attack}$ is the weight of the attacker.

$\theta_{defense}$ is the weight of the defender.

\setlength{\parindent}{0ex}
NB: The optimization problem under constraint is a not zero-sum game.
\\
\subsection{Motivation}
Another representation of the problem which could seem more intuitive is to minimize/maximize, zero-sum game, the final error for the defender/attacker while the adversarial examples are not too distant to the original image (constraint).

\begin{equation}\label{eq:minmax}
\max_{\theta_{attack}} \min_{\theta_{defense}} E_{x \sim \chi}(L_{\alpha}(x, Defense(Attack(x, \theta_{attack}), \theta_{defense})))
\end{equation}

Under the constraint:
\begin{equation}\label{eq:constraint2}
E_{x \sim \chi}(L_{\alpha}(Attack(x, \theta_{attack}), x)) \leq \beta
\end{equation}

\paragraph{}
The first representation (Equations \ref{eq:defense}, \ref{eq:attack}, \ref{eq:constraint}) is preferred to the second (Equations \ref{eq:minmax}, \ref{eq:constraint2}) for two reasons:
\begin{itemize}
\item The attacks against neural networks try to minimize the perturbation of the input, while the second representation does not. 
\item The problem optimization is harder to solve because the perturbations generated by state of art attacks like DeepFool \cite{moosavi-dezfooli_deepfool:_2015} are not on the constraint border (when $\beta$ is large enough to accept all types of perturbations).
\end{itemize}

\begin{figure}[H]
\centering
\subfloat[First representation]{\includegraphics[scale=0.5]{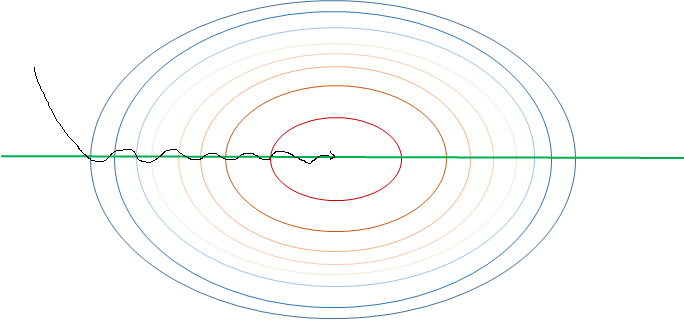}\label{repr:f1}}
\subfloat[Second representation]{\includegraphics[scale=0.5]{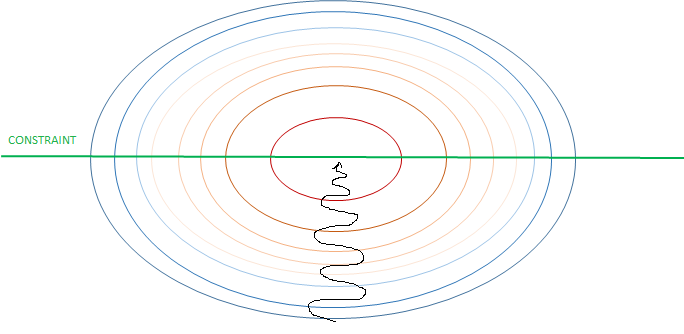}\label{repr:f2}}
\caption{Two representations of the optimization problem.}
\label{repr}
\end{figure}

During the training, the optimization process will follow these strategies (See figure \ref{repr}).
\begin{itemize}
\item In the first representation: at the begining, the optimizer will reach the border of the constraint and then it will follow it until reaching the optimum.
\item In the second representation, the optimizer will explore the space of solution while continuously closing the gap with the border, reaching it only at the end of the training.
\end{itemize}

This is the second reason why the first representation is preferred to the second: the constraint border is easier to follow.
To solve this under constraint optimization problem, we choose to add the constraint in the training loss:

\begin{equation}\label{eq:loss}
\begin{multlined}
\min_{\theta_{attack}} E_{x \sim \chi}(L_{\alpha}(Attack(x, \theta_{attack}), x)) + \\ \gamma * \max( E_{x \sim \chi}(L_{\alpha}(x, Defense(Attack(x, \theta_{attack}), \theta_{defense}))) - \beta, 0.0)
\end{multlined}
\end{equation}

\paragraph{}
Where $\gamma$ is big enough to satisfy the constraint. This representation differs from the classic Lagrangian because we want the constraint gradient part to be free when the constraint is satisfied. This approximation of the under constraint optimization problem is justified by the fact that the inequality loss should be equal or close to $\beta$.

\subsection{Model and training procedure}
\paragraph{}
The attacker is implemented as an auto-encoder where a uniform random vector and the features of the encoder are concatenated and then passed through the generator. Adding a uniform random vector will allow the attacker to generated noise without modifying the "real" features. The defender is implemented as a classic auto-encoder. In a first phase, to initialize the weights of the networks, the two auto-encoders are trained to generate realistic images (the constraint is ignored, see Equation [\ref{eq:defense}] and [\ref{eq:attack}]).

\paragraph{}
In the second phase, the weights of the Attacker‘s encoder and the Defender‘s generator are fixed. This choice is justified by:

\begin{itemize}
\item Attacker‘s encoder: The goal of the attacker is to generate adversarial examples from given non perturbed examples. Because image information is compressed in the features, the attacker could be summarized by a generator which has the features generated by the encoder as inputs, already learned, and a random noise.
\item Defender‘s generator: The goal of the defender is to be robust against adversarial attacks. This means that the features of adversarial examples should be identical to the features of non-perturbed examples. This is done by fixing the generator parameters after the initialization phase, in this manner the features of perturbed input will be generated in the same way as normal inputs.
\end{itemize}

\paragraph{}
During this second phase we optimize the equations [\ref{eq:defense}] and [\ref{eq:loss}].

\subsection{Testing procedure}
As auto-encoder is an architecture made for unsupervised learning, no classifier is used during training. In order to be able to used adversarial attacks and evaluate the network robustness a classifier is trained with the defender’s features generated by the encoder with data coming from the dataset, not the adversarial data generated by the attacker. The robustness of the classifier is evaluated against the followings attacks: DeepFool $L_{2}$ and $L_{\infty}$, Single Pixel, LocalSearch, FGSM and BIM. Robustness is evaluated by two criteria, the attack success rate and the noise level of adversarial examples. Noise level is defined by:
\begin{equation}
Noise_{i} = \sqrt{\sum_{i}(X_{i} - Adv_{i})^{2}}
\end{equation}
\\
Where $X_{i}$ is the original image and $Adv_{i}$ is the generated adversarial example.

\section{MNIST Experiments}
\subsection{Configuration}
This method has been evaluated on the MNIST dataset with the following networks:

\begin{figure}[H]
\centering
\subfloat[Attacker architecture]{\includegraphics[scale=0.5]{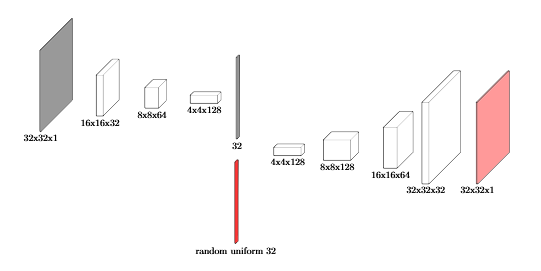}\label{archi:f1}}
\subfloat[Defender architecture]{\includegraphics[scale=0.5]{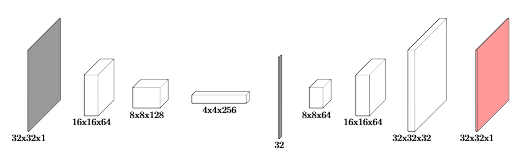}\label{archi:f2}}
\label{architecture}
\caption{Network architecures}
\end{figure}

\paragraph{}
The neural network use convolutions with kernel size 3x3, strides 2x2 and ReLU activation function. The feature layers are not passed through an activation function. The last layers have $(1+ \tanh)/2$ as activation function. The batch size is equal to 128. 

The Adam optimizer is used with the following learning rates:
\begin{itemize}
\item Defender : Initialization phase = $5 * 10^{-4}$, Optimization phase = $10^{-5}$
\item Attacker: Initialization phase = $10^{-3}$, Optimization phase = $10^{-5}$
\end{itemize}

The constants were fixed at $\gamma$ = 5.0, $\beta$ = 0.01, the $L_{2}$ loss is used.
The initialization has been done during 9 epochs and the optimization phase during 31 epochs.
Then a classifier is trained during epochs 20 with Adam (learning rate $10^{-3}$, loss: softmax-cross-entropy). The classifier uses the Defender’s encoder (whose weights are fixed) and a hidden dense layer of 64 units with ReLU activation.
NB: The networks have not been optimized for this task and no search was done for the optimal parameters  $\gamma$ and $\beta$. However, several learning rates have been tested to achieve the presented results: a bad choice of learning rate could imply a non-robust network.

\subsection{Results}
\paragraph{Adversarial attacks tools}
Tools have been developped to facilitate adversarial attacks implementation such as Cleverhans\cite{papernot2018cleverhans} or Foolbox \cite{rauber_foolbox:_2017}. Those tools are open-source Frameworkds that propose state-of-art algorithms used to generate adversarial examples on any model.

\paragraph{}
In order to compare our method against a traditional network, we trained a classic autoencoder, with an architecure similar to Defender architectrure in \Cref{archi:f2}. We then train a classifier as  previously descripted on each network. After 20 epochs, the classifier achieves an accuracy around 97$\%$ on MNIST test set.
Then, the network with classifier is attacked by several methods on the whole test set, using Foolbox \cite{rauber_foolbox:_2017}

\begin{table}[H]
\centering
\begin{tabular}{|l|l|l|}
\hline
Attack Name & Classic Auto Encoder & Our Method \\
\hline
DeepFool (iter=50 – $L_{2}$) & 94,41 $\%$ / 1,25 & \: \textbf{0.54 $\%$ / 0.0075}\\
DeepFool (iter=50 – $L_{\infty}$)  & \:\: 100 $\%$ / 1,81 & \:  \textbf{4.13 $\%$ / 1.13} \\
FGSM ($\epsilon$=0.005) & \: 0.41 $\%$ / 0.07 &  \textbf{1.61 $\%$ / 0.04} \\
Iterative ($\epsilon$=0.005) & \: 1.04 $\%$ / 0.08 & \textbf{1.25 $\%$ / 0.04} \\
Single Pixel & \: 9.17 $\%$ / 1.0 &  \:  \textbf{4.72 $\%$ / 1.0} \\
LocalSearch & 44.41 $\%$ / 5.22 & \textbf{27.51 $\%$ / 9.71} \\
\hline
\end{tabular}
\caption{Attack results : Attack Success Rate / Noise Level} 
\end{table}

* The evaluation is done only on the 1000 first image of the dataset, due to computation time. \\
**The noise level is the mean of the $L_{2}$ norm between the original image and the adversarial image for all successful attacks.

\begin{table}[H]
\centering
\begin{tabular}{|l|l|l|}
\hline
Attack Name & MagNet & Our Method \\
\hline
DeepFool (iter=50 – $L_{\infty}$) & \textbf{0.6} $\%$ & \:  4.13 $\%$ \\
FGSM ($\epsilon$=0.005) & 3.2 $\%$ & \textbf{1.61} $\%$ \\
Iterative ($\epsilon$=0.005) & 4.8 $\%$ & \textbf{1.25} $\%$ \\
\hline
\end{tabular}
\caption{MagNet comparaison} 
\end{table}

\begin{center}
\begin{tikzpicture}[scale=1]
\begin{axis}[axis x line=bottom, axis y line = left, xlabel={Iterations}, ylabel={Success Rate}, title={DeepFool attacks}, ymode=log, legend entries={FUNN - $L_{2}$ norm, FUNN - $L_{\inf}$ norm, Classic AE - $L_{2}$ norm, Classic AE - $L_{\inf}$ norm}, legend style={at={(1.03,1)},anchor=north west}, legend plot pos=right]
\addplot coordinates{(50, 0.54) (100, 0.75) (200, 1.35)};
\addplot coordinates{(50, 4.13) (100, 39.55) (200, 89.20)};
\addplot coordinates{(50, 94.41) (100, 99.93) (200, 100)};
\addplot coordinates{(50, 100) (100, 100) (200, 100)};
\end{axis}
\end{tikzpicture}
\end{center}

\subsection{Analysis}
\paragraph{DeepFool with $L_{2}$ norm}
In the previous table we can see that only 0.5 $\%$ of attack succeeded when DeepFool was used with 50 iterations (number of iterations used by the original article of DeepFool) for a noise level mean around 0.0075. That was expected, as during learning, the attacker tries to find noise level greater than 0.01 (in fact, “~equal to” like explained above) which can fool the defender. If the defender succeeded to defend against this level of noise, we expect to have as good a defense against any level of noise lower than this threshold. The classifier seems to be robust against $L_{2}$ DeepFool attacks for 50 iterations.
When the number of deepfool iterations increases, the success rate of the attack slowly increases: for 200 iterations, 1.35$\%$ of attack succeeded. It’s only when the number of iterations reaches 1000 steps that the percentage of successful attacks becomes very significant (15.25$\%$). However, the noise increases significantly at the same time to reach 0.73 which is far from 0.01.
We may think that increasing the threshold would allow to increase the robustness but in the case of MNIST it is impossible because doing so, the attacker converges to a black image. Indeed, because the $L_{2}$ distance between a MNIST image and a black image is around 0.07, it is not possible to increase the threshold too much.  

\paragraph{DeepFool with $L_{\infty}$ norm}
With the default step used by DeepFool (50 steps), the classifier seems to be robust (4.13$\%$ of success rate) for the $L_{\infty}$ norm. However, the classifier is not robust when the number of steps increases: at 200 steps, the success rate of the attacks reaches 89.20$\%$. This is the reason why it is important to test DeepFool with several steps which is unfortunately not often done in the defense literature. However, even with 50 steps, the norm of the noise is strong with 1.13. This is  the reason why it does not necessary mean that the network is not robust at all against this attack.

\paragraph{Single Pixel and LocalSearch}
Even if the number of success of the attack is relatively low (4.72$\%$), this level of success is higher than what we may expect. In our case this attack looks more powerful than DeepFool which is designed precisely to minimize the $L_{2}$ norm. We would except a result lower than with DeepFool.

\paragraph{Gradient Methods}
FGSM and Gradient have a high attack success rate, close to 100$\%$ for Gradient Sign and Gradient. This result was expected because the defender is not trained with this high level of noise, shown by all these methods.

\paragraph{MagNet Comparaison}
Our method have similar performance to MagNet \cite{meng_magnet:_2017} without using any adversarial detectors that induces more computation time during inference.

\section{Conclusion}
\paragraph{}
Without adding extra computation complexity, like filtering the image or detecting adversarial images, we demonstrated that a classic network can be robust against $L_{2}$ attacks by an adapted unsupervised training procedure. This procedure shows promising results toward robust unsupervised networks. However, even if it was expected due to the design of the defense, the trained network is not robust against all types of attacks such as gradient attacks. Further studies should be focused on a more general defense against different attacks with unsupervised training. We will also focus our future work on adapting this type of learning procedure to real world datasets with more complex network architectures.

\newpage
\bibliography{biblio}{}

\begin{thebibliography}{10}

\bibitem{baluja_learning_nodate}
Shumeet Baluja and Ian Fischer.
\newblock Learning to {Attack}: {Adversarial} {Transformation} {Networks}.
\newblock page~9.

\bibitem{carlini_defensive_2016}
Nicholas Carlini and David Wagner.
\newblock Defensive {Distillation} is {Not} {Robust} to {Adversarial}
  {Examples}.
\newblock {\em arXiv:1607.04311 [cs]}, July 2016.
\newblock arXiv: 1607.04311.

\bibitem{folz_adversarial_2018}
Joachim Folz, Sebastian Palacio, Joern Hees, Damian Borth, and Andreas Dengel.
\newblock Adversarial {Defense} based on {Structure}-to-{Signal}
  {Autoencoders}.
\newblock {\em arXiv:1803.07994 [cs, stat]}, March 2018.
\newblock arXiv: 1803.07994.

\bibitem{goodfellow_explaining_2014}
Ian~J. Goodfellow, Jonathon Shlens, and Christian Szegedy.
\newblock Explaining and {Harnessing} {Adversarial} {Examples}.
\newblock {\em arXiv:1412.6572 [cs, stat]}, December 2014.
\newblock arXiv: 1412.6572.

\bibitem{hinton_distilling_2015}
Geoffrey Hinton, Oriol Vinyals, and Jeff Dean.
\newblock Distilling the {Knowledge} in a {Neural} {Network}.
\newblock {\em arXiv:1503.02531 [cs, stat]}, March 2015.
\newblock arXiv: 1503.02531.

\bibitem{kurakin_adversarial_2016-1}
Alexey Kurakin, Ian Goodfellow, and Samy Bengio.
\newblock Adversarial examples in the physical world.
\newblock {\em arXiv:1607.02533 [cs, stat]}, July 2016.
\newblock arXiv: 1607.02533.

\bibitem{kurakin_adversarial_2016}
Alexey Kurakin, Ian Goodfellow, and Samy Bengio.
\newblock Adversarial {Machine} {Learning} at {Scale}.
\newblock {\em arXiv:1611.01236 [cs, stat]}, November 2016.
\newblock arXiv: 1611.01236.

\bibitem{lee_generative_2017}
Hyeungill Lee, Sungyeob Han, and Jungwoo Lee.
\newblock Generative {Adversarial} {Trainer}: {Defense} to {Adversarial}
  {Perturbations} with {GAN}.
\newblock {\em arXiv:1705.03387 [cs, stat]}, May 2017.
\newblock arXiv: 1705.03387.

\bibitem{meng_magnet:_2017}
Dongyu Meng and Hao Chen.
\newblock {MagNet}: {A} {Two}-{Pronged} {Defense} against {Adversarial}
  {Examples}.
\newblock In {\em Proceedings of the 2017 {ACM} {SIGSAC} {Conference} on
  {Computer} and {Communications} {Security} - {CCS} '17}, pages 135--147,
  Dallas, Texas, USA, 2017. ACM Press.

\bibitem{metzen_detecting_2017}
Jan~Hendrik Metzen, Tim Genewein, Volker Fischer, and Bastian Bischoff.
\newblock On {Detecting} {Adversarial} {Perturbations}.
\newblock {\em arXiv:1702.04267 [cs, stat]}, February 2017.
\newblock arXiv: 1702.04267.

\bibitem{moosavi-dezfooli_deepfool:_2015}
Seyed-Mohsen Moosavi-Dezfooli, Alhussein Fawzi, and Pascal Frossard.
\newblock {DeepFool}: a simple and accurate method to fool deep neural
  networks.
\newblock {\em arXiv:1511.04599 [cs]}, November 2015.
\newblock arXiv: 1511.04599.

\bibitem{munusamy_kabilan_vectordefense:_2018}
Vishaal Munusamy~Kabilan, Brandon Morris, and Anh Nguyen.
\newblock {VectorDefense}: {Vectorization} as a {Defense} to {Adversarial}
  {Examples}.
\newblock April 2018.

\bibitem{narodytska_simple_2016}
Nina Narodytska and Shiva~Prasad Kasiviswanathan.
\newblock Simple {Black}-{Box} {Adversarial} {Perturbations} for {Deep}
  {Networks}.
\newblock {\em arXiv:1612.06299 [cs, stat]}, December 2016.
\newblock arXiv: 1612.06299.

\bibitem{papernot2018cleverhans}
Nicolas Papernot, Fartash Faghri, Nicholas Carlini, Ian Goodfellow, Reuben
  Feinman, Alexey Kurakin, Cihang Xie, Yash Sharma, Tom Brown, Aurko Roy,
  Alexander Matyasko, Vahid Behzadan, Karen Hambardzumyan, Zhishuai Zhang,
  Yi-Lin Juang, Zhi Li, Ryan Sheatsley, Abhibhav Garg, Jonathan Uesato, Willi
  Gierke, Yinpeng Dong, David Berthelot, Paul Hendricks, Jonas Rauber, and
  Rujun Long.
\newblock Technical report on the cleverhans v2.1.0 adversarial examples
  library.
\newblock {\em arXiv preprint arXiv:1610.00768}, 2018.

\bibitem{papernot_distillation_2015}
Nicolas Papernot, Patrick McDaniel, Xi~Wu, Somesh Jha, and Ananthram Swami.
\newblock Distillation as a {Defense} to {Adversarial} {Perturbations} against
  {Deep} {Neural} {Networks}.
\newblock {\em arXiv:1511.04508 [cs, stat]}, November 2015.
\newblock arXiv: 1511.04508.

\bibitem{rauber_foolbox:_2017}
Jonas Rauber, Wieland Brendel, and Matthias Bethge.
\newblock Foolbox: {A} {Python} toolbox to benchmark the robustness of machine
  learning models.
\newblock {\em arXiv:1707.04131 [cs, stat]}, July 2017.
\newblock arXiv: 1707.04131.

\bibitem{samangouei_defense-gan:_2018}
Pouya Samangouei, Maya Kabkab, and Rama Chellappa.
\newblock {DEFENSE}-{GAN}: {PROTECTING} {CLASSIFIERS} {AGAINST} {ADVERSARIAL}
  {ATTACKS} {USING} {GENERATIVE} {MODELS}.
\newblock page~17, 2018.

\bibitem{santhanam_defending_2018}
Gokula~Krishnan Santhanam and Paulina Grnarova.
\newblock Defending {Against} {Adversarial} {Attacks} by {Leveraging} an
  {Entire} {GAN}.
\newblock {\em arXiv:1805.10652 [cs, stat]}, May 2018.
\newblock arXiv: 1805.10652.

\bibitem{shen_ape-gan:_2017}
Shiwei Shen, Guoqing Jin, Ke~Gao, and Yongdong Zhang.
\newblock {APE}-{GAN}: {Adversarial} {Perturbation} {Elimination} with {GAN}.
\newblock {\em arXiv:1707.05474 [cs]}, July 2017.
\newblock arXiv: 1707.05474.

\bibitem{srinivasan_counterstrike:_2018}
Vignesh Srinivasan, Arturo Marban, Klaus-Robert Müller, Wojciech Samek, and
  Shinichi Nakajima.
\newblock Counterstrike: {Defending} {Deep} {Learning} {Architectures}
  {Against} {Adversarial} {Samples} by {Langevin} {Dynamics} with {Supervised}
  {Denoising} {Autoencoder}.
\newblock {\em arXiv:1805.12017 [cs, stat]}, May 2018.
\newblock arXiv: 1805.12017.

\bibitem{szegedy_intriguing_2013}
Christian Szegedy, Wojciech Zaremba, Ilya Sutskever, Joan Bruna, Dumitru Erhan,
  Ian Goodfellow, and Rob Fergus.
\newblock Intriguing properties of neural networks.
\newblock {\em arXiv:1312.6199 [cs]}, December 2013.
\newblock arXiv: 1312.6199.

\bibitem{yuan_adversarial_2017}
Xiaoyong Yuan, Pan He, Qile Zhu, and Xiaolin Li.
\newblock Adversarial {Examples}: {Attacks} and {Defenses} for {Deep}
  {Learning}.
\newblock {\em arXiv:1712.07107 [cs, stat]}, December 2017.
\newblock arXiv: 1712.07107.

\end{thebibliography}
\bibliographystyle{plain}
\end{document}